\documentclass[12pt,a4paper]{article}
\usepackage[margin=25mm]{geometry}
\usepackage{amsmath}
\usepackage{amsfonts}
\usepackage{amssymb}
\usepackage{graphicx}
\usepackage{lineno} 
\usepackage{booktabs}
\usepackage[utf8]{inputenc}
\usepackage{authblk}
\usepackage{setspace}
\usepackage{algorithm}
\usepackage{algorithmic}
\usepackage{url}
\providecommand{\keywords}[1]{\textbf{\textit{Keywords: }} #1}

\begin{document}

\title{Coronavirus Optimization Algorithm: A bioinspired metaheuristic based on the COVID-19 propagation model}

\author{F. Martínez-Álvarez$^1$\thanks{Corresponding author, fmaralv@upo.es}, G. Asencio-Cortés$^1$, J. F. Torres$^1$,\\D. Gutiérrez-Avilés$^1$, L. Melgar-García$^1$, R. Pérez-Chacón$^1$, \\C. Rubio-Escudero$^2$, J. C. Riquelme$^2$, A. Troncoso$^1$}

\affil{$^1$Data Science \& Big Data Lab, Pablo de Olavide University, ES-41013 Seville, Spain}
\affil{$^2$Department of Computer Science, University of Seville, ES-41012 Seville, Spain }
\date{}

\doublespacing
\maketitle

\begin{abstract}
A novel bioinspired metaheuristic is proposed in this work, simulating how the coronavirus spreads and infects healthy people. From an initial individual (the patient zero), the coronavirus infects new patients at known rates, creating new populations of infected people. Every individual can either die or infect and, afterwards, be sent to the recovered population. Relevant terms such as re-infection probability, super-spreading rate or traveling rate are introduced in the model in order to simulate as accurately as possible the coronavirus activity. The Coronavirus Optimization Algorithm has two major advantages compared to other similar strategies. First, the input parameters are already set according to the disease statistics, preventing researchers from initializing them with arbitrary values. Second, the approach has the ability of ending after several iterations, without setting this value either. Infected population initially grows at an exponential rate but after some iterations, when considering social isolation measures and the high number recovered and dead people, the number of infected people starts decreasing in subsequent iterations. Furthermore, a parallel multi-virus version is proposed in which several coronavirus strains evolve over time and explore wider search space areas in less iterations. Finally, the metaheuristic has been combined with deep learning models, in order to find optimal hyperparameters during the training phase. As application case, the problem of electricity load time series forecasting has been addressed, showing quite remarkable performance.
\end{abstract}


\section{Introduction}
The coronavirus (COVID-19) is a new respiratory virus, firstly discovered in humans in December 2019, that has spread worldwide, having been reported more than 1 million infected people so far. Much remains unknown about the virus, including how many people may have very mild or asymptomatic infections, and whether they can transmit the virus. The precise dimensions of the outbreak are hard to know.

Bioinspired models typically mimic behaviors from the nature and are known for their successful application in hybrid approaches to find parameters in machine learning model optimization. Viruses can infect people and these people can either die, infect other people or simply get recovered after the disease. Vaccines and the immune defense system typically fight the disease and help to mitigate their effects while an individual is still infected. This behavior is typically modeled by an $SIR$ model, consisting of three kind of individuals: $S$ for the number of susceptible, $I$ for the number of infectious, and $R$ for the number of recovered.


Metaheuristics must deal with huge search spaces, even infinite for the continuous cases, and must find suboptimal solutions in reasonable execution times. The rapid propagation of the coronavirus along with its ability of infecting most of the countries in the world impressively fast, has inspired the novel metaheuristic proposed in this work, named Coronavirus Optimization Algorithm (CVOA). A parallel version is also proposed in order to spread different coronavirus strains and achieve better results in less iterations.  

The main CVOA advantages regarding other similar approaches can be summarized as follows: 
\begin{enumerate}
\item Coronavirus statistics are known by the scientific community. In this sense, the rate of infection, the mortality rate or the re-infection probability are already known. That is, CVOA is parametrized with actual values for rates and probabilities, preventing the user to perform an additional study on the most suitable setup configuration.

\item CVOA can stop the solutions exploration after several iterations, with no need to be configured. That is, the number of infected people increases during the first iterations, however, after a certain number of iterations, the number of infected people starts decreasing, until reaching a void infected set of individuals.

\item The coronavirus high spreading rate is useful for exploring promising regions more thoroughly (intensification) while the use of parallel strains ensures that all regions of the search space are evenly explored (diversification).

\item Another relevant contribution of this work is the proposal of a new codification, discrete and of dynamic length, specifically designed for combining Long Short-Term Memory networks (LSTM) with CVOA (or any other metaheuristic). 
\end{enumerate}

As for the limitations of the current approach, there is mainly one. Since there is no vaccine currently, it has not been included in the procedure to reduce the number of individuals candidates to be infected. This fact involves an exponential increase of the infected population in the first iterations and, therefore, an exponential increase of the execution time for such iterations. This fact, however, is partially solved with the social isolation measures that simulate individuals that cannot be infected at a particular iteration. 

A study case is included in this work to discuss the CVOA performance. CVOA has been used to find the optimal values for the hyperparameters of a LSTM architecture \cite{KELOTRA20}, which is widely used model for artificial recurrent neural network (RNN), in the field of deep learning \cite{CNUDDE19}. Data from the Spanish electricity consumption have been used to validate the accuracy. The results achieved verge on 0.45\%, substantially outperforming other well-established methods such as random forest, gradient-boost trees, linear regression or deep learning optimized with other metaheuristics. The code, developed in Phyton with a discrete codification, is available in the supplementary material (along with an academic version in Java for a binary codification).

Finally, it is acknowledged the need of further study on the performance of well-known functions \cite{GLOVER03}, however, given the relevance of coronavirus is acquiring throughout the world (declared as pandemic by the World Health Organization) and the remarkable results achieved when combined with deep learning, it was wanted to share this work hoping it inspires future research in this direction. 

The rest of the paper is organized as follows. Section \ref{sec:related} discusses related and recent works. The methodology proposed is introduced in Section \ref{sec:methodology}. Section \ref{sec:hybridization} proposes a discrete codification to hybridize deep learning models with CVOA and provides some illustrative cases. A preliminary analysis on how populations are created and evolved over time is discussed in Section \ref{sec:preliminary}. The results achieved are reported and discussed in Section \ref{sec:results}. Finally, the conclusions drawn and future work suggestions are included in Section \ref{sec:conclusions}. 

\section{Related works}\label{sec:related}
There are many bioinspired metaheuristics to solve optimization problems. Although CVOA has been conceived to optimize any kind of problems, this section focuses on optimization algorithms applied to hybridize deep learning models. 

It is hard to find consensus among the researchers on which method should be applied to which problem, and, for this reason, many optimization methods have been proposed during the last decade to improve deep learning models. Generally, the criterion for selecting a method is its associated performance from a wide variety of perspectives. Low computation cost, accuracy or even implementation difficulty can be accepted as one of these criteria.

The Virus Optimization Algorithm (VOA) was proposed by Liang and Cuevas-Juárez in 2016 \cite{LIANG16} and later improved in \cite{LIANG20}. However, as many other metaheuristics, the results of its application are highly dependent on its initial configuration. Additionally, it simulates generic viruses, without adding individualized properties for particular viruses. The results achieved indicate that its usefulness is beyond doubt. 

One of the most extended metaheuristics used to improve deep learning parameters is genetic algorithms (GA). Hence, a LSTM network optimized with GA can be found in \cite{CHUNG18}. To evaluate the proposed hybrid approach, the daily Korea Stock Price Index data were used, outperforming the benchmark model. In 2019, a network traffic prediction model based on LSTM and GA was proposed in \cite{CHEN19}. The results were compared to pure LSTM and ARIMA, reporting higher accuracy.

Multi-agents systems have also been applied to optimize deep learning models. The use of Particle Swarm Optimization (PSO) can be found in \cite{LIU19}. The authors proposed a model based on kernel principal component analysis and back propagation neural network with PSO for midterm power load forecasting. The hybridization of deep learning models with PSO was also explored in \cite{JUNIOR19} but, this time, the authors applied the methodology with image classification purposes. 

Ants colony optimization (ACO) models have also been used to hybridize deep learning. Thus, Desell et al. \cite{DESELL15} proposed an evolving deep recurrent neural networks using ACO applied to the challenging task of predicting general aviation flight data. The work in \cite{ELSAID18} introduced a method based on ACO to optimize a LSTM recurrent neural networks. Again, the field of application was flight data records obtained from an airline containing flights that suffered from excessive vibration.

Some papers exploring the Cuckoo Search (CS) properties have been published recently as well. In \cite{SRIVASTAVA19}, CS was used to find suitable heuristics for adjusting the hyper-parameters of another LSTM network. The authors claimed an accuracy superior to 96\% for all the datasets examined. Nawi et al. \cite{NAWI14} proposed the use of CS to improve the training of RNN in order to achieve fast convergence and high accuracy. Results obtained outperformed those than other metaheuristics. 

The use of the artificial bee colony (ABC) optimization algorithm applied to LSTM can also be found in the literature. Hence, and optimized LSTM with ABC to forecast the bitcoin price was introduced in \cite{YULIYONO19}. The combination of ABC and RNN was also proposed in \cite{BOSIRE19} for traffic volume forecasting. This time the results were compared to standard backpropagation models. 

From the analysis of these works, it can be concluded that there is an increasing interest in using metaheuristics in LSTM models. However, not as many works as for artificial neural networks can be found in the literature and, none of them, based on a virus propagation model. These two facts, among others, justify the application of CVOA to optimize LSTM models. 

\section{Methodology}\label{sec:methodology}
This section introduces the CVOA methodology. Thus, Section \ref{sec:steps} describes the steps for a single strain. Section \ref{sec:multi-agent} introduces the modifications added to use CVOA as a parallel version. Section \ref{sec:parameters} suggests how parameters must be set. Section \ref{sec:pseudocodes} shows the CVOA pseudo codes and comments them. 

\subsection{Steps}\label{sec:steps}
\textbf{Step 1.} Generation of the initial population. The initial population consists of one individual, the so-called patient-zero ($PZ$). As in the coronavirus epidemic, it identifies the first human being infected.

\noindent \textbf{Step 2.} Disease propagation. Depending on the individual, several cases are evaluated: 
\begin{enumerate}
    
\item Some of the infected individuals die. They die according to the coronavirus death rate ($P\_DIE$). Such individuals can no longer infect new individuals. 
\item The individuals surviving the coronavirus will infect new individuals (intensification). Two types of spreading are considered, according to a given probability ($P\_SUPERSPREADER$): 
\begin{itemize}

\item Ordinary spreaders. Infected individuals will infect new ones according to the coronavirus spreading rate ($SPREADING\_RATE$). 
\item Super-spreaders. Infected individuals will infect new ones according to the coronavirus superspreading rate ($SUPERSPREADING\_RATE$).
\end{itemize}
\item There is another consideration, since it is needed to ensure diversification. Both ordinary and super-spreaders individuals can travel and explore solutions quite dissimilar. Therefore, individuals have a probability of traveling ($P\_TRAVEL$) thus allowing to propagate the disease to solutions that may be quite different ($TRAVELER\_RATE$). In case of not being traveler, new solutions will change according to an $ORDINARY\_RATE$. Note that one individual can be both super-spreader and traveler.
\end{enumerate}

\noindent \textbf{Step 3.} Updating populations. Three populations are maintained and updated for each generation.
\begin{enumerate}
\item	Dead population. If any individual dies, it is added to this population and can never be used again. 
\item	Recovered population. After each iteration, infected individuals (after spreading the coronavirus according to the previous step) are sent to the recovered population. It is known that there is a reinfection probability. Hence, an individual belonging to this population could be re-infected at any iteration provided that it meets the reinfection criterion ($P\_REINFECTION$). Another situation must be considered, since individuals can be isolated simulating they are implementing the social distancing measures. For the sake of simplicity, it is considered that an isolated individual is sent to the recovered population as well when meeting an isolation probability ($P\_ISOLATION$). 
\item New infected population. This population gathers all individuals infected at each iteration, according the procedure described in the previous steps. It is possible that repeated new infected individuals are created at each iteration and, consequently, it is recommended to remove such repeated individuals from this population before the next iteration starts running. 
\end{enumerate}

\noindent \textbf{Step 4.} Stop criterion. One of the most interesting features of the proposed approach lies on its ability to end without the need of controlling any parameter. This situation occurs because the recovered and dead populations are constantly growing as time goes by, and the new infected population cannot infect new individuals. It is expected that the number of infected individuals increases for a certain number of iterations. However, from a particular iteration on, the size of the new infected population will be smaller than that of the current one because recovered and dead populations are too big, and the size of the infected population decays over time. Additionally, a preset number of iterations ($PANDEMIC\_DURATION$) can be added to the stop criterion. The social isolation measures also contributes to reaching the stop criterion.

\subsection{Remarks for a parallel CVOA version}\label{sec:multi-agent}
It must be noted that it is very simple to use CVOA in a multi-virus version since it can be implemented as a population-based algorithm, when considering the pandemic as a set of intelligent agents each of them evolving in parallel. In contrast to trajectory-based metaheuristics, population-based focuses on the diversification in the search space. 

For this case, a new variable must be defined, $strains$, which will determine the number of strains that will be launched in parallel. Each strain could simulate different regions. In other words, strains can be differently configured so that each of them intensifies with their own rates. 

Several considerations must be done for this case: 
\begin{enumerate}
    \item Every strain is run independently, following the steps in the previous section.
    \item A wise strategy should be followed to generate $PZs$ for each strain. For instance, it is suggested the generation of orthogonal $PZs$ or with high Hamming distances. That way, a wider search space could be covered, enhancing diversification. 
    \item The interaction between the different strains is done by means of dead and recovered populations, which must be shared by all the strains. Operations over these populations must be handled as concurrent updates \cite{DHAR19}.
    \item New infected populations, on the contrary, are different for each strain and no concurrent operations are required.
    \item This version may help to simulate different rates for different strains. That way, if there is any initial information about the search space, some strains could be more focused on diversification and some others on intensification. 
\end{enumerate}

Depending on the hardware resources and how busy they are, every strain may evolve at different speeds. This situation should not pose any problems since it is known that the pandemic evolves at different rates and starts at different time stamps depending on region of the world. 

Last, another application can be found for this parallel version. CVOA emulates an SIR model and consequently, any other global pandemic could be modeled by using the known rates of, for instance, the flu of 1918 or 1957, another coronavirus SARS or MERS, HIV, or Ebola. Hypothetically, parallel pandemics could be run with different rates. 

\subsection{Suggested parameters setup}\label{sec:parameters}
Since CVOA simulates the coronavirus disease propagation, most of the rates (propagation, re-infection or mortality) are already known. This fact prevents the research from wasting time in selecting values for such rates and turns the CVOA into metaheuristic quite easy to execute. 

However, it must be noted that the current rates are not definitive yet and it is expected they will vary over time, as the pandemic evolves. Maybe these values will not be stable until 2021 or even 2022. The suggested values have been retrieved from the World Health Organization \cite{WHO} and are discussed below:

\begin{enumerate}
\item $P\_DIE$. An infected individual can die with a given probability. Currently, this rate is set as almost 5\% by the scientific community. Therefore, $P\_DIE = 0.05$. 

\item $P\_SUPERSPREADER$. It is the probability that an individual spread the disease to a greater number of healthy individuals. It is known that this situation affects to a 10\% of the population, therefore, $P\_SUPERSPREADER = 0.1$. After this condition is validated, two situations can be found: 

\begin{itemize}
    
\item $ORDINARY\_RATE$. If the infected individual is not a super-spreader, then the infection rate (also known as reproductive number, $R_0$) is 2.5. It is suggested that this rate varies from 0 to 5. 
\item $SUPERSPREADER\_RATE$. If the infected individual turns out to be super-spreader, then he/she infects up to 15 healthy individuals on average. It is suggested that this rate varies from 6 to 15. 
\end{itemize}

\item $P\_REINFECTION$. It is known that a recovered individual can be re-infected. The current reported rate is 14\%. Therefore, $P\_REINFECTION = 0.14$.

\item $P\_ISOLATION$. This value is uncertain because countries are taking different measures for social isolation. This parameter helps to reduce the exponential growth of the infected population after each iteration. Therefore, a high value must be assigned to this probability. It is suggested that $P\_ISOLATION = 0.5$.

\item $P\_TRAVEL$. This probability simulates how an infected individual can travel to any place in the world and can infect healthy individuals. It is known that almost a 10\% of the population travel during a week (simulated time for every iteration), so $P\_TRAVEL = 0.1$. 

\item $PANDEMIC\_DURATION$. This parameter simulates the duration of the pandemic. Since the estimated recovering time is one week, each iteration simulates one week. Currently, this data is unknown so this number can be adjusted to the size of the problem. It is suggested that $PANDEMIC\_DURATION = 30$.

\item $strains$. This parameter should be adjusted according to the size of the problem and the hardware availability, and it is difficult to suggest a value suitable for all situations. But a tentative initial value could be five, in an attempt to simulate one different strain per continent. Therefore, $strains=5$. Another important decision that must be made is how to initialize every $PZ$ associated with the strains. When just one strain is considered, $PZ$ is suggested to be randomly initialized. However, with $strains > 1$ the user should search for orthogonal $PZs$ and to uniformly distribute them in the search space. This strategy should help to cover bigger search spaces in less iterations and to evaluate individuals with maximal distances. 
\end{enumerate}

\subsection{Pseudo codes}\label{sec:pseudocodes}
This section provides the pseudo code of the most relevant functions for the CVOA, along with some comments to better understand them. 

\subsubsection{Function $CVOA$}
This is the main function and its pseudo code can be found in Algorithm \ref{alg:main}. Four lists must be maintained: dead, recovered, infected (the current set of infected individuals) and new infected individuals (the set of new infected individuals, generated by the spreading of the coronavirus from the current infected individuals).

The initial population is generated by means of the patient zero ($PZ$), which is a random solution. 

The number of iterations is controlled by the main loop, evaluating the duration of the pandemic (preset value) and if there is still any infected individual. In this loop, every individual can either die (it is sent to the dead list) or infect, thus enlarging the size of the new infected population. How this infection is made, is implemented in function $infect$ (see Section \ref{sec:infect}). 

Once the new population is formed, qll individuals are evaluated and if any of them outperforms the best current one, the latter is updated. 

\begin{algorithm}
\caption{Function $\mathbf{cvoa}$}\label{alg:main}
\begin{algorithmic}[1]
\STATE \textbf{define} infectedPopulation, newInfectedPopulation \textbf{as} $set$ \textbf{of} $Individual$
\STATE \textbf{define} dead, recovered  \textbf{as} $list$ \textbf{of} $Individual$
\STATE \textbf{define} PZ, bestIndividual, currentBestIndividual, aux  \textbf{as} $Individual$
\STATE \textbf{define} time \textbf{as} $integer$
\STATE \textbf{define} bestSolutionFitness, currentbestFitness \textbf{as} $real$
\STATE time $\leftarrow$ 0
\STATE PZ $\leftarrow$ InfectPatientZero()
\STATE infectedPopulation  $\leftarrow$ PZ
\STATE bestIndividual $\leftarrow$ PZ
\WHILE{time $<$ $EPIDEMIC\_DURATION$ $AND$ \textbf{sizeof}(infectedPopulation) $>$ 0 }
    \STATE dead  $\leftarrow$ die(infectedPopulation)
    \FORALL{$i \in infectedPopulation$}
        \STATE aux  $\leftarrow$ infect(i,recovered,dead)
        \IF{\textbf{notnull}(aux)}
        \STATE newInfectedPopulation  $\leftarrow$ aux
        \ENDIF
    \ENDFOR
    \STATE currentBestIndividual $\leftarrow$ selectBestIndividual(newInfectedPopulation)
    \IF{fitness(currentBestIndividual) $>$ bestIndividual}
    \STATE bestIndividual $\leftarrow$ currentBestIndividual
    \ENDIF
    \STATE recovered  $\leftarrow$ infectedPopulation
    \STATE \textbf{clear}(infectedPopulation)
    \STATE infectedPopulation $\leftarrow$ newInfectedPopulation
    \STATE time $\leftarrow$ time $+$ 1
\ENDWHILE
\RETURN bestIndividual
\end{algorithmic}
\end{algorithm}

\subsubsection{Function $infect$}\label{sec:infect}
This function receives an infected individual and returns the set of new infected individuals. Two additional lists, recovered and dead, are also received as input parameters since they must be updated after the evaluation of every infected individuals. The pseudo code is shown in Algorithm \ref{alg:infect}.

Two conditions are evaluated to determine the number of new infected individuals (use of $SPREADER\_RATE$ or $SUPERSPREADER\_RATE$) or how different the new individuals will be ($ORDINARY\_RATE$ or $TRAVELER\_RATE$. The implementation on how these new infected individuals are encoded according to such rates is carried out in the function $newInfection$.

\begin{algorithm}
\caption{Function $\mathbf{infect}$}\label{alg:infect}
\begin{algorithmic}[1]
\REQUIRE infected \textbf{as} \textbf{of} $Individual$; recovered, dead \textbf{as} $list$ \textbf{of} $Individual$
\STATE \textbf{define} R1, R2 \textbf{as} $real$
\STATE \textbf{define} newInfected \textbf{as} $list$ \textbf{of} $Individual$
\STATE R1 $\leftarrow$ RandomNumber()
\STATE R2 $\leftarrow$ RandomNumber()
\IF{R1 $<$ $P\_TRAVEL$} 
    \IF{R2 $<$ $P\_SUPERSPREADER$} 
        \STATE newInfected $\leftarrow$ newInfection (infected, recovered, dead, $SPREADER\_RATE$, $ORDINARY\_RATE$)
    \ELSE
    \STATE newInfected $\leftarrow$ newInfection (infected, recovered, dead, $SUPERSPREADER\_RATE$, $ORDINARY\_RATE$)
    \ENDIF
\ELSE
    \IF{R2 $<$ $P\_SUPERSPREADER$} 
        \STATE newInfected $\leftarrow$ newInfection (infected, recovered, dead, $SPREADER\_RATE$, $TRAVELER\_RATE$)
    \ELSE
        \STATE newInfected $\leftarrow$ newInfection (infected, recovered, dead, $SUPERSPREADER\_RATE$, $TRAVELER\_RATE$)
    \ENDIF
\ENDIF
\RETURN newInfected
\end{algorithmic}
\end{algorithm}

\subsubsection{Function $newInfection$}
Given an infected individual this function generates new infected individuals, according to the spreading and traveling rates. This function also controls that the new infected individuals are not already in the dead list (in such case this new infection is ignored) or in the recovered list (in such case the $P\_REINFECTION$ is applied to determine whether the individual is re-infected or if it remains in the recovered list). Additionally, it considers that the new potential infected individual might be isolated, which is controlled by $P\_ISOLATION$. Although the use of an extra list could be implemented, it has been decided to treat these individuals as recovered. Therefore, if an isolated individual is attempted to be infected, it is added to the recovered list. 

The effective generation of the new infected individuals must be carried in the function $replicate$, whose pseudo code is not provided because it depends on the codification and the nature of the problem to be optimized. This function must return a set of new infected individuals, according to the aforementioned rates. Specific information on how this codification and replication is done for LSTM models.

The pseudo code for the described procedure can be found in Algorithm \ref{alg:newInfection}.

\begin{algorithm}
\caption{Function $\mathbf{newInfection}$}\label{alg:newInfection}
\begin{algorithmic}[1]
\REQUIRE infected \textbf{as} $Individual$; recovered, dead \textbf{as} list \textbf{of} $Individual$
\STATE \textbf{define} R3, R4 \textbf{as} $real$
\STATE \textbf{define} newInfected \textbf{as} $list$ \textbf{of} $Individual$
\STATE R3 $\leftarrow$ RandomNumber()
\STATE R4 $\leftarrow$ RandomNumber()
\STATE aux $\leftarrow$ replicate(infected, $SPREAD\_RATE$, $TRAVELER\_RATE$)
\FORALL{$i \in aux$}
    \IF{i $\not\in$ dead}
        \IF{i $\not\in$ recovered} 
            \IF{$R4 > P\_ISOLATION$}
            \STATE newInfected $\leftarrow$ i
            \ELSE
            \STATE $recovered \leftarrow i$
            \ENDIF
        \ELSIF{$R3 < P\_REINFECTION$}
            \STATE newInfected $\leftarrow$ i
            \STATE \textbf{remove} i \textbf{from} recovered
        \ENDIF
    \ENDIF
\ENDFOR
\RETURN newInfected
\end{algorithmic}
\end{algorithm}

\subsubsection{Function $die$}
This function is called from the $main$ function. It evaluates all individuals in the infected population and determines whether they die or not, according to the given $P_DIE$. Those meeting this condition, are sent to the dead list. Algorithm \ref{alg:die} describes this procedure.

\begin{algorithm}
\caption{Function $\mathbf{die}$}\label{alg:die}
\begin{algorithmic}[1]
\REQUIRE infectedPopulation \textbf{as} $list$ \textbf{of} $Individual$
\STATE \textbf{define} dead \textbf{as} $list$ \textbf{of} $Individual$
\STATE \textbf{define} R5 \textbf{as} $real$
\FORALL{i $\in$ infectedPopulation} 
    \STATE R5 $\leftarrow$ RandomNumber()
    \IF{R5 $<$ $P\_DIE$} 
        \STATE dead $\leftarrow$ i
    \ENDIF
\ENDFOR
\RETURN dead
\end{algorithmic}
\end{algorithm}

\subsubsection{Function $selectBestIndividual$}
This is an auxiliary function used to find the best fitness in a list of infected individuals. Its peudo code is shown in Algorithm \ref{alg:best}.

\begin{algorithm}
\caption{Function $\mathbf{selectBestIndividual}$}\label{alg:best}
\begin{algorithmic}[1]
\REQUIRE infectedPopulation \textbf{as} $list$ \textbf{of} $Individual$
\STATE \textbf{define} bestIndividual \textbf{as} $Individual$
\STATE \textbf{define} bestFitness \textbf{as} $real$
\STATE bestFitness $\leftarrow$ $MINVALUE$
\FORALL{i $\in$ infectedPopulation} 
    \IF{fitness(i) $>$ bestFitness} 
        \STATE bestFitness $\leftarrow$ fitness(i)
        \STATE bestIndividual $\leftarrow$ i
    \ENDIF
\ENDFOR
\RETURN bestIndividual
\end{algorithmic}
\end{algorithm}

\section{Hybridizing deep learning with CVOA}\label{sec:hybridization}

This section describes the codification proposed for an individual, in order to hybridize deep learning with CVOA. The term hybridize is used in this context as the combination of two computational techniques (deep learning and CVOA) so that the best hyperparameter values are discovered. This strategy is very common in machine learning for optimizing models during the training process \cite{CALVET17,DARWISH20, DEVIKANNIGA19}.

\begin{figure}[h!]
    \centering
    \includegraphics[width=\textwidth]{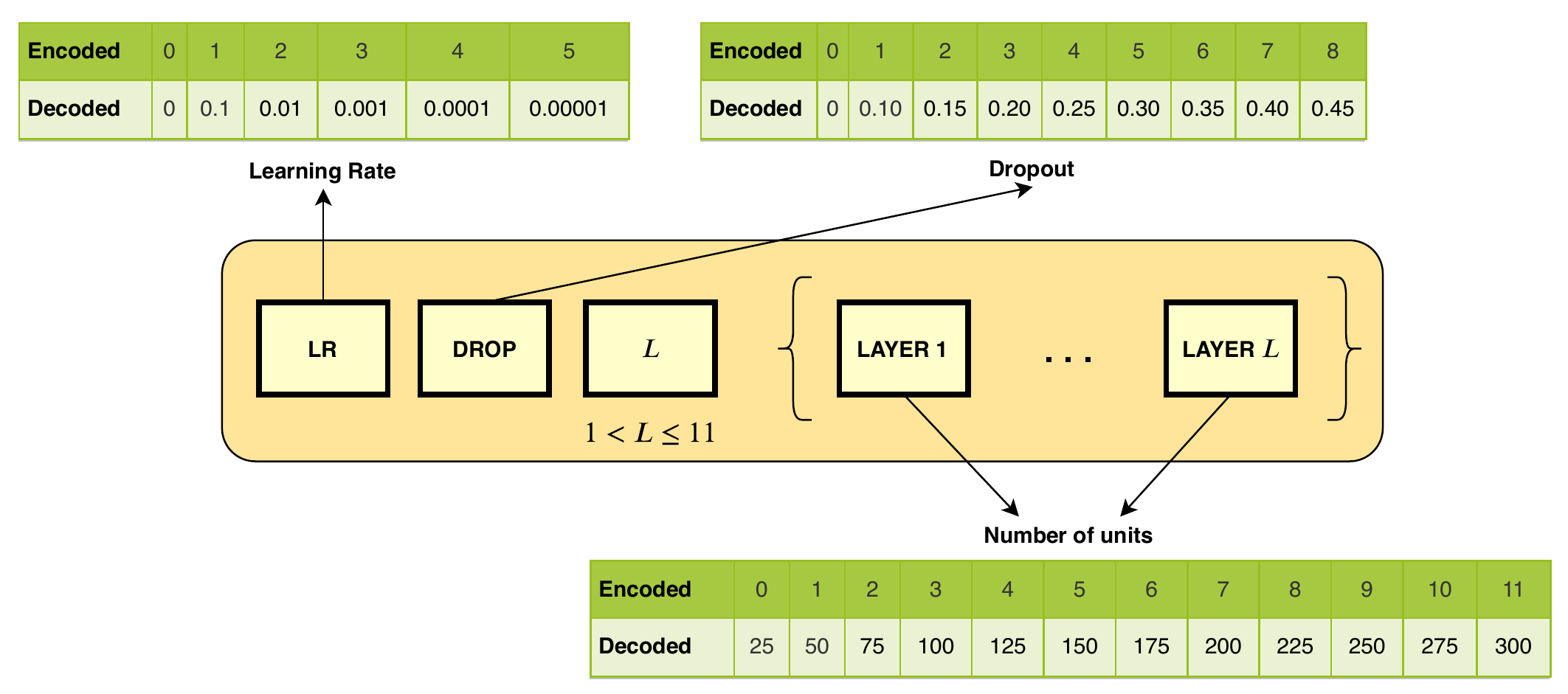}
    \caption{Individual codification for hybridizing deep learning architectures using the proposed CVOA algorithm.}
    \label{fig:individual-codification}
\end{figure}

Hence, the individual codification shown in Figure \ref{fig:individual-codification} has been implemented in order to apply CVOA to optimize deep neural network architectures.

As it can be seen in Figure \ref{fig:individual-codification}, each individual is composed of the following elements. The element LR encodes the learning rate used in the neural network algorithm. It can take a value from $0$ to $5$ and its corresponding decoded values are $0$, $0.1$, $0.01$, $0.001$, $0.0001$ and $0.00001$. 

The element DROP encodes the dropout rate applied to the neural network. It can take values from $0$ to $8$ that correspond to $0$, $0.10$, $0.15$, $0.20$, $0.25$, $0.30$, $0.35$, $0.40$ and $0.45$, respectively. The dropout rate is distributed uniformly for all the layers of the network. That is, if the dropout is $0.4$ and the network has $4$ layers, then the $10\%$ ($0.1$) of the neurons of each layer will be removed.

The element $L$ of the individual stores the number of layers of the network. It is restricted to $1 < L \leq 11$. The first layer is referred to the input layer of the neural network. The rest of layers are hidden layers. The output layer is excluded from the codification. Therefore, the optimized network can contain from $1$ to $10$ hidden layers.

The proposed individual codification has a variable size. Thus, its size depends on the number of layers indicated in the element $L$. Consequently, a list of elements (LAYER $1$, ..., LAYER $L$) are also included in the individual, which encode the number of units (neurons) for each network layer. Each of these elements can take values from $0$ to $11$, and their corresponding decoded values range from $25$ to $300$, with a step of $25$.

\subsection{PZ generation}

The PZ, as it has been described previously, is the individual of the first iteration in the CVOA algorithm. Following the hybridization proposed, a random individual is created considering the codification defined above.

In first place, a random value for the learning rate of the PZ is generated. Specifically, a number between $0$ and $5$ is generated randomly in a uniform distribution. Such limits are indicated in Figure \ref{fig:individual-codification}, according to the possible encoded values of the learning rate element. The same process is carried out to produce a random value for the dropout element. In such case, a random number between $0$ and $8$ is generated.

In second place, a random number of layers is generated for the element $L$ of $PZ$. Such number of layers is a random number between $2$ and $11$. Note that the first layer is reserved for the input layer of the neural network, as it has been discussed before.

In last place, for each one of the $L$ layers, a random number of units is generated between $0$ and $11$, covering the possible encoded values for the number of units previously defined (see Figure \ref{fig:individual-codification}).

\subsection{Infection procedure}\label{sec:infectProc}

The infection procedure described here corresponds to the functionality of $replicate()$, introduced in the line $4$ of the Algorithm \ref{alg:newInfection}. This procedure takes an individual as input and returns an infected individual according to the following procedure.

The first step is to determine the element $L$ of the infected individual that will be mutated. The probability of such mutation occurs has been set to $\frac{1}{3}$ so that every element has the same probability to mutate. If the mutation occurs, then the element $L$ of the individual is modified according to the process described in Section \ref{sec:results-single-pos-mut}.

If the element $L$ (the number of layers of the network) changes, then the elements encoding the different layers within the individual (LAYER $1$, ..., LAYER $L$) must be resized accordingly. Such resizing process is explained in Section \ref{sec:results-resizing}.

The second step is to determine how many elements of the individual will be infected. If the $TRAVELER\_RATE < 0$, then the number of infected elements is generated randomly from $0$ to the length of the individual (excluding the element $L$). Else, the $TRAVELER\_RATE$ indicates itself the number of infected elements.

As third step, once it is determined the number of infected elements of the individual, a list of random positions is generated. For example, if three positions of the individual must be changed, then the random positions affected could be, for instance, whose referred to the elements \{DROP, LAYER $2$, LAYER $4$\}.

Finally, the selected positions of the individual are mutated. Such mutation is described in Section \ref{sec:results-single-pos-mut}.

\subsection{Individual resizing process}
\label{sec:results-resizing}

When an individual is infected at the position of the element $L$, the list of elements that encodes the number of units per layer (LAYER $1$, ..., LAYER $L$) must be resized accordingly.

In the case that the new number of layers after the infection is lower than its previous value, then the last leftover elements are removed. For instance, if the initial individual is $\{2,0,4\}\{3,2,1,6\}$ (four layers), the element $L=4$ is infected and the new value is $L=2$, then the resulting individual will be $\{2,0,2\}\{3,2\}$.

In the case that the new number of layers after the infection is higher than its previous value, the new random elements are added at the end of the individual. For instance, if the initial individual is $\{2,0,4\}\{3,2,1,6\}$ (four layers), the element $L=4$ is infected and the new value is $L=6$, then the resulting individual could be $\{2,0,6\}\{3,2,1,6,0,4\}$.

\subsection{Single position mutation}
\label{sec:results-single-pos-mut}

The process carried out to change the value of a specific element of an individual is described in this section. 

First, a signed change amount $C \in \{-2,-1,+1,+2\}$ is randomly determined using the following criteria. A random real number $P$ between $0$ and $1$ is generated using a uniform distribution. If $P<0.25$, then the change amount will be $C=-2$. Else if $P<0.5$, then the change amount will be $C=-1$. Else if $P<0.75$, then the change amount will be $C=+1$. Else, the change amount will be $C=+2$.

Once the amount of change is determined, the new value for the infected element is computed. If its previous value is $V$, then the new value after the single position mutation will be $V'=V+C$. If the new value $V'$ exceeds the limits defined for the individual codification, such value is set to the maximum or minimum allowed value accordingly.

\section{CVOA preliminary analysis}\label{sec:preliminary}
This section provides an overview on how populations evolve over time, and how the search space is explored to reach the optimum value for a given fitness function.

To conduct this experimentation, a simple binary codification has been used. The fitness function was $f(x) = (x-15)^2$ because to goal of this section is to evaluate the growth of the new infected populations, and not to the find challenging optimum values. This function reaches the minimum value at $x=15$, that is, $f(15)=0$.

For this reason, individuals with 10, 20, 30, 40 and 50 bits have been tested. Tables \ref{tab:10bit}-\ref{tab:50bit} summarize the results achieved for each of these lengths, respectively. Every experiment has been launched 50 times, determining that, on average, the optimum value was found for 11, 12, 14, 15 and 17 iterations, respectively. Each table shows the results of an execution meeting this criterion.

\begin{table}[!ht]
    \centering
    \caption{Sample execution for a 10-bit binary codification.\label{tab:10bit}}
    \begin{tabular}{ccccc}
    \hline
    Iteration &	Deaths	& Recovered &	Infected &	Fitness \\
    \hline
    1&	0&	1&	6&	7.72E+04\\
    2&	1&	7&	8&	6.05E+04\\
    3&	1&	15&	5&	6,00E+04\\
    4&	1&	20&	6&	4\\
    5&	2&	26&	7&	4\\
    6&	2&	33&	5&	4\\
    7&	2&	38&	19&	4\\
    8&	3&	48&	13&	1\\
    9&	3&	61&	16&	1\\
    10&	4&	77&	18&	1\\
    11&	5&	95&	20&	0\\
    \hline
    \end{tabular}
\end{table}

\begin{table}[!ht]
    \centering
    \caption{Sample execution for a 20-bit binary codification.\label{tab:20bit}}
    \begin{tabular}{ccccc}
    \hline
    Iteration &	Deaths	& Recovered &	Infected &	Fitness \\
    \hline
    1&0&	1&	10&	5.73E+11\\
    2&1&	11&	14&	3.92E+11\\
    3&2&	25&	16&	3.14E+11\\
    4&2&	41&	24&	1.33E+09\\
    5&4&	63&	36&	1.37E+07\\
    6&5&	98&	56&	2.76E+06\\
    7&8&	153&72&	1.32E+06\\
    8&12&222&112&1.61E+04\\
    9&17&330&172&1\\
    10&26&499&249&1\\
    11&38&738&389&1\\
    12&58&1112&567&0\\
    \hline
    \end{tabular}
\end{table}

\begin{table}[!ht]
    \centering
    \caption{Sample execution for a 30-bit binary codification.\label{tab:30bit}}
    \begin{tabular}{ccccc}
    \hline
    Iteration &	Deaths	& Recovered &	Infected &	Fitness \\
    \hline
    1& 0& 	1& 	8& 	6.11E+17\\
    2&1& 	9& 	9& 	6.00E+16\\
    3&1& 	18& 	18& 	5.59E+16\\
    4&2&  	36& 	21& 	1.12E+16\\
    5&3&  57& 	43&  	1.04E+16\\
    6&5& 	100& 	83& 	1.85E+13\\
    7&9& 	183& 	126& 	1.08E+10\\
    8&16& 	307& 	208& 	1.51E+09\\
    9&26& 	514& 	326& 	1.21E+09\\
    10&42& 	838& 	606& 	1.07E+09\\
    11&73& 	1436& 	1049& 	1.69E+07\\
    12&125& 	2467& 	1796& 	676\\
    13&215& 	4233& 	2967& 	9\\
    14&363& 	7138& 	4823& 	0\\
    \hline
    \end{tabular}
\end{table}

\begin{table}[!ht]
    \centering
    \caption{Sample execution for a 40-bit binary codification.\label{tab:40bit}}
    \begin{tabular}{ccccc}
    \hline
    Iteration &	Deaths	& Recovered &	Infected &	Fitness \\
    \hline
    1 &	0 &	1 &	10 &	2.23E+18\\
    2 &	1 &	11 &	20 &	2.57E+15\\
    3 &	2 &	31 &	41 &	1.83E+11\\
    4 &	4 &	70 &	85 &	3.58E+09\\
    5 &	8 &	154	 &170 &	3.24E+09\\
    6 &	17 &	321	 &331 &	7.39E+08\\
    7 &	33 &	652 &	598 &	7.08E+08\\
    8 &	63	 &1246 &	1036 &	1.06E+08\\
    9 &	115 &	2275	 &1918	 &2.12E+07\\
    10 &	211 &	4169 &	3592 &	1612900\\
    11 &	390 &	7716 &	6357 &	270400\\
    12 &	708 &	14003 &	11205 &	2809\\
    13 &	1268 &	25052 &	19495 &	49\\
    14 &	2243 &	44198 &	33215 &	9\\
    15 &	3904 &	76672 &	55122 &	0\\
    \hline
    \end{tabular}
\end{table}

\begin{table}[!ht]
    \centering
    \caption{Sample execution for a 50-bit binary codification.\label{tab:50bit}}
    \begin{tabular}{ccccc}
    \hline
    Iteration &	Deaths	& Recovered &	Infected &	Fitness \\
    \hline
    1	&0	&	1	&	12	&	4.61E+18\\
    2	&	1	&	13	&	23	&	7.80E+15\\
    3	&2	&	36	&	23	&	3.62E+13\\
    4	&	2	&	36	&	40	&	1.39E+12\\
    5	&	4	&	76	&	71	&	1.37E+12\\
    6	&	8	&	146	&	119	&	1.57E+10\\
    7	&	14	&	265	&	185	&	7.50E+08\\
    8	&	23	&	449	&	338	&	1.21E+08\\
    9	&	40	&	787	&	586	&	7963684\\
    10	&	69	&	1369&	1100&	7873636\\
    11	&	124	&	2461	&	2129	&	597529\\
    12	&	230	&	4579	&	3957	&	68121\\
    13	&	428	&	8499	&	7211	&	17956\\
    14	&	789	&	15644	&	13305	&	36\\
    15	&	1454	&	28807	&	24167	&	36\\
    16	&	2662	&	52622	&	43184	&	1\\
    17	&	4821	&	95116	&	76288	&	0\\

    \hline
    \end{tabular}
\end{table}

Table \ref{tab:search} summarizes the amount of search space explored, on average, before finding the optimum value. For a small space of $2^{10}=1024$ possible values, the optimum one is reached after exploring 15.6250\% valid solutions. However, this value acutely decreases as the search space increases. The opposite case is reached for $2^{50} = 1.1259E+15$ possible values, where the optimum value was reached after exploring just 0.00000002\% of valid solutions. 

\begin{table}[!ht]
    \centering
    \caption{Search space explored for a binary codification with different individuals length.\label{tab:search}}
    \begin{tabular}{cccc}
    \hline
   	Length & Search space &	Evaluated &	Evaluated (\%)\\
    \hline
    10 &	1024&	160	&15.62500000\%\\
    20&	1048576	&2253&	0.21486282\%\\
    30&	1073741824&	15589&	0.00145184\%\\
    40&	1,09951E+12&	170827&	0.00001554\%\\
    50&	1,1259E+15&	219613&	0.00000002\%\\
    \hline
    \end{tabular}
\end{table}

\begin{figure}
    \centering
    \includegraphics[width=\textwidth]{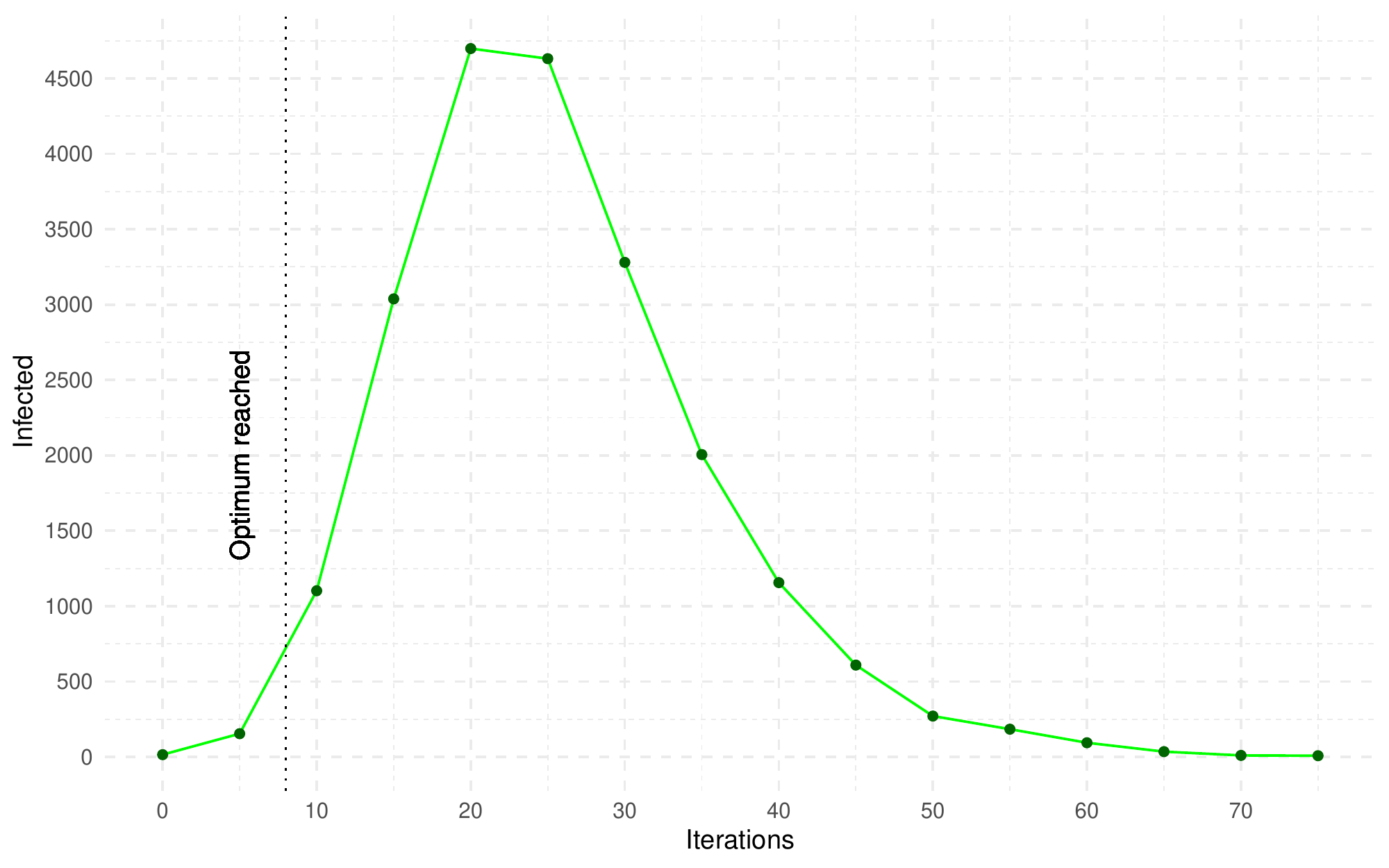}
    \caption{Number of new infected individuals for a 20-bit binary codification execution.}
    \label{fig:Infected}
\end{figure}

Figure \ref{fig:Infected} illustrates how the new infected population evolves over time for the 20-bit codification case with the suggested parameters initialization. The number of new people increases until a given iteration in which it starts to decrease. Note that data shown in the figure are sampled every five iterations. 

\begin{figure}
    \centering
    \includegraphics[width=\textwidth]{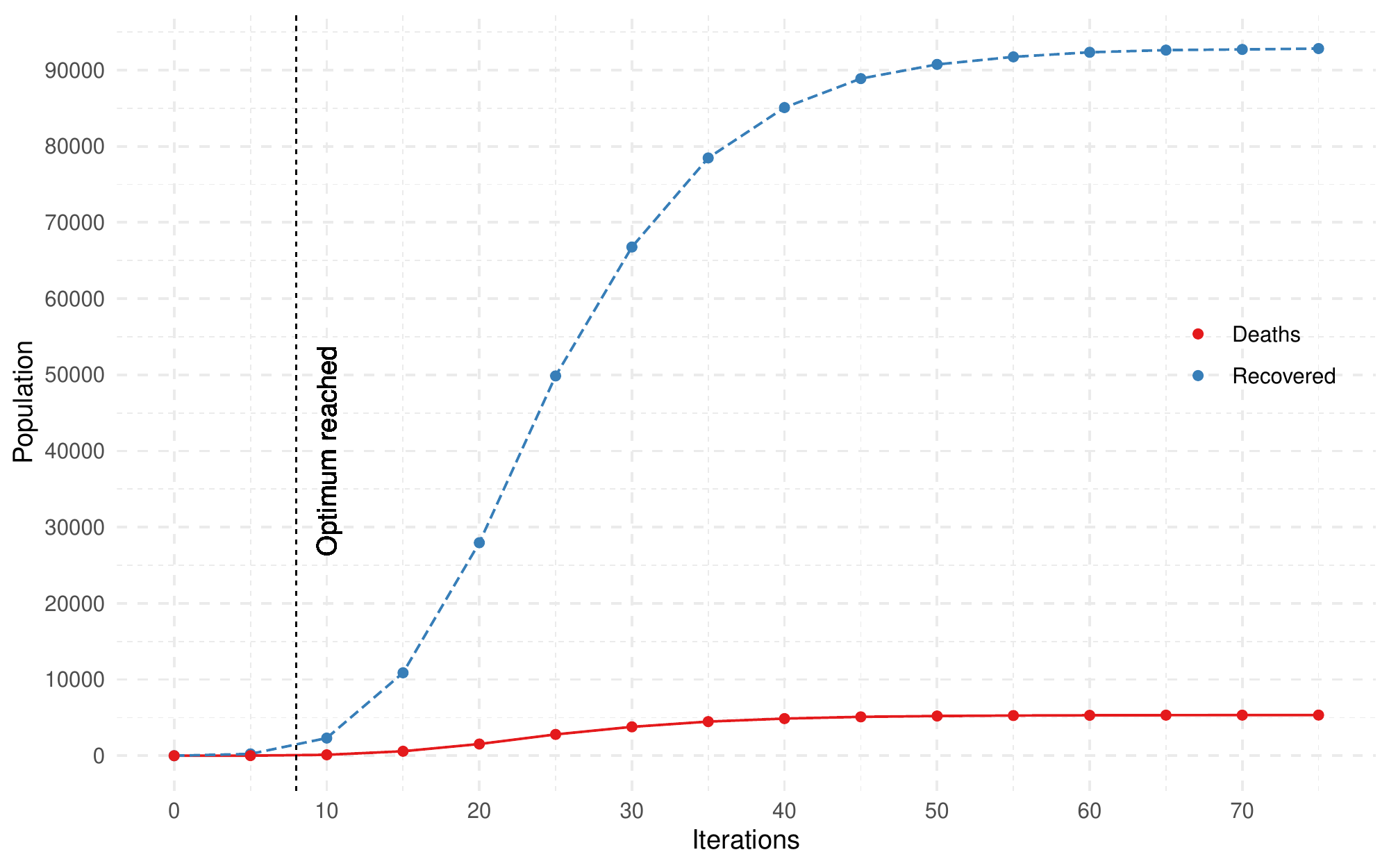}
    \caption{Total number of recover and dead people for a 20-bit binary codification execution.}
    \label{fig:recover-dead}
\end{figure}

Finally, Figure \ref{fig:recover-dead} shows the number of recovered and dead people. These two curves accumulate these numbers since dead and recovered people are sent to their respective lists and are no longer infected (except for those in recovered that can be reinfected for a its given probability $P\_REINFECTION$). The final number of recovered is 92826 and of dead people 5335, being 94.56\% and 5.44\%, respectively. 

\section{Results}\label{sec:results}
This section reports the results achieved by hybridizing a deep learning model with CVOA. Section \ref{sec:results-study-case} describes the study case selected to prove the effectiveness of the proposed algorithm. Section \ref{sec:dataset} describes the dataset used. Section \ref{sec:performance} discusses the results achieved and includes some comparative methods. 

\subsection{Study case: electricity demand time series forecasting}
\label{sec:results-study-case}

The forecasting of future values fascinates the human being. To be able to understand how certain variables evolve over time has many benefits in many fields. 

Electricity demand forecasting is not an exception, since there is a real need for planning the amount to be generated or, in some countries, to be bought. 

The use of machine learning to forecast such time series has been intensive during the last years \cite{MARTINEZ15}. But, with the development of deep learning models, and, in particular of LSTM, much research is being conducted in this application field \cite{BEDI19}.  

\subsection{Dataset description}\label{sec:dataset}
The time series considered in this study is related to the electricity consumption in Spain from January 2007 to June 2016, the same as used in \cite{TORRES18}. It is a time series composed of 9 years and 6 months with a 10-minute sampling frequency, resulting in 497832 measures. 

As in the original paper, the prediction horizon is 24, that is, this is a multi-step strategy with $h=24$. The size of samples used for the prediction of these 24 values is 168. Furthermore, the dataset was split into 70\% for the training set and 30\% for the test set, and in addition, a 30\% of the training set has also been selected for the validation set, in order to find the optimal parameters. The training set covers the period from January 1, 2007 at 00:00 to August 20, 2013 at 02:40. Therefore, the test set comprises the period from August 20, 2013 at 02:50 to June 21, 2016 at 23:40.

\subsection{Performance analysis}\label{sec:performance}
This section reports the results obtained by hybridizing LSTM with CVOA, by means of the codification proposed in Section \ref{sec:hybridization}, to forecast the Spanish electricity dataset described in Section \ref{sec:dataset}. 

Linear regression (LR), decision tree (DT), gradient-boosted trees (GBT) and random forest (RF) models have been used with a parametrization setups according to those studied in \cite{GALICIA17,GALICIA19}. A deep neural network optimized with a grid search (DNN-GS) according to \cite{TORRES18} has also been applied. Another deep neural network, but optimized with random search (DNN-RS) and smoothed with a low-pass filter (DNN-RS-LP) \cite{TORRES19}, has also been applied. Furthermore, CVOA has been combined with DNN (DNN-CVOA), using the same codification as in LSTM.

These results along with those of LSTM, and combinations with GS, RS, RS-LP and CVOA are summarized in Table \ref{tab:CVOA-LSTM}, expressed in terms of the mean absolute percentage error (MAPE). It can be observed that LSTM-CVOA outperforms all evaluated methods which have showed particularly remarkable performance for this real-world dataset. Additionally, DNN-CVOA outperforms all other DNN configurations which confirms the superiority of CVOA with reference to GS, RS, and RS-LP. 

Another relevant consideration that must be taken into account is that the compared methods generated 24 independent models, each of them for every value forming $h$. So, it would expected that LSTM-CVOA performance increases if independent models are generated for each of the values in $h$.  

\begin{table}[!h]
\caption{Results in terms of MAPE for CVOA-LSTM compared to other well established methods.}\label{tab:CVOA-LSTM}
\centering
\begin{tabular}{lc}
\hline
Method	&	MAPE (\%)	\\ 
\hline
LR		    &	7.34	\\
DT		    &	2.88	\\
GBT	        &	2.72	\\
RF		    &	2.20	\\ 
DNN-GS	    &	1.68	\\
DNN-RS      &   1.57    \\ 
DNN-RS-LP   &   1.36    \\
DNN-CVOA    &   1.18    \\
LSTM-GS	    &	1.22	\\
LSTM-RS     &   0.84    \\ 
LSTM-RS-LP  &   0.82    \\
\textbf{LSTM-CVOA} &  \textbf{0.47}    \\ 
\hline
\end{tabular}
\end{table}

These results have been achieved with the individual $\{4,0,8\}\{9,7,2,7,2,7,10,7\}$, which decoded involves the following architecture parameters: 
\begin{itemize}
    \item Learning rate: 10E-04.
    \item Dropout: 0.
    \item Number of layers: 8. 
    \item Units per layer: $[250, 200, 75, 200, 75, 200, 275, 200]$
\end{itemize}

\begin{figure}[!h]
    \centering
    \includegraphics[width=\textwidth]{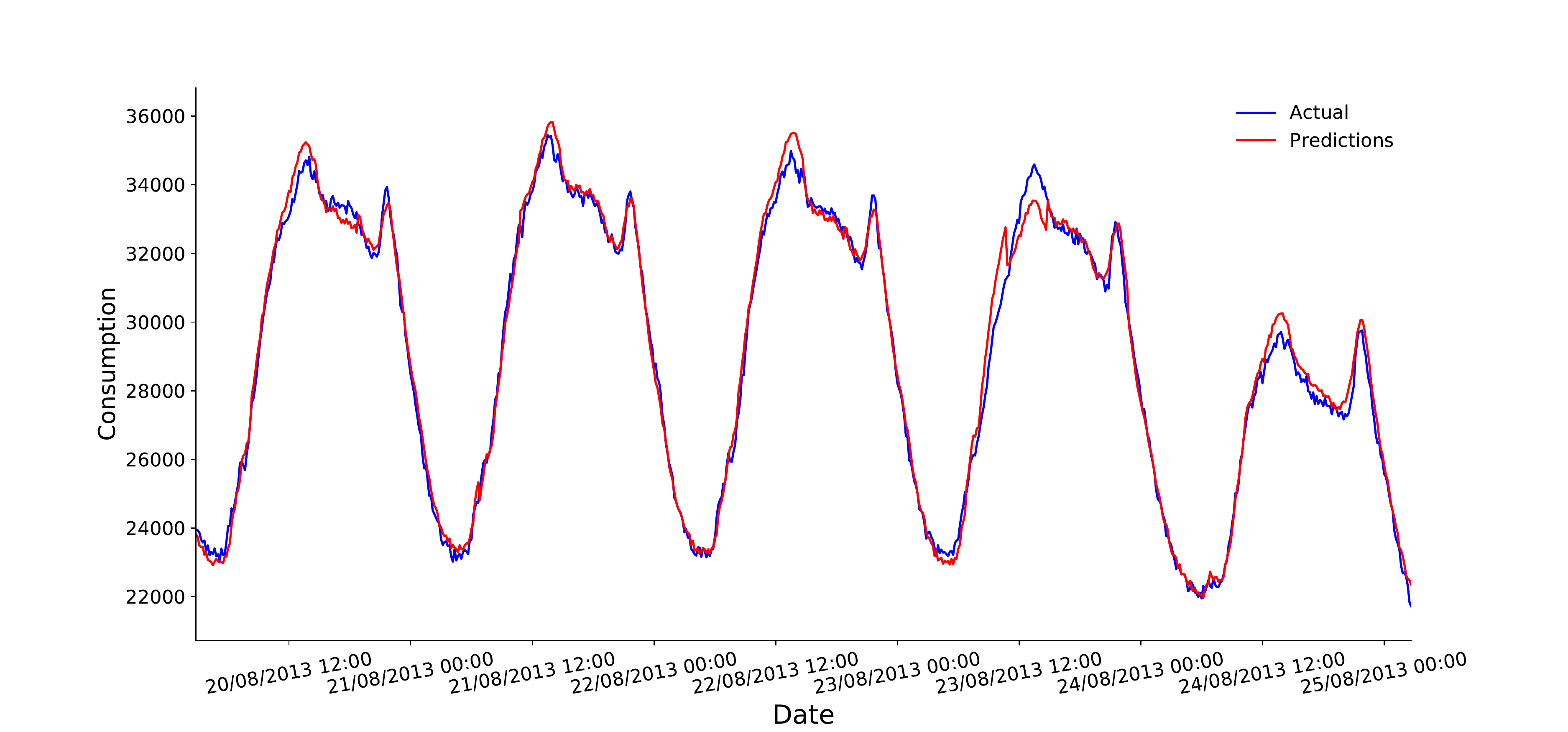}
    \caption{Actual versus predicted values for the first five days in the test set (in watts).}
    \label{fig:comparison}
\end{figure}

Finally, Figure \ref{fig:comparison} depicts the first five predicted days versus their actual values, expressed in watts.

\section{Conclusions and future works}\label{sec:conclusions}
This work has introduced a novel bioinspired metaheuristic, based on the coronavirus behavior. On the one hand, CVOA has two major advantages. First, its highly relation to the coronavirus spreading model, prevents the authors to make any decision about the inputs' values. Second, it ends after a certain number of iterations due to the exchange of individuals between healthy and dead/recovered lists. Additionally, a novel discrete and dynamic codification has been proposed to hybridize deep learning models. On the other hand, it exhibits some limitations. Such is the case for the exponential growth of the infected population as time (iterations) goes by.

Furthermore, a parallel version is proposed so that CVOA is easily transformed into a multi-virus metaheuristic, in which different coronavirus strains search for the best solution in a collaborative way. This fact allows to model every strain with different initial setups (higher $DEATH\_RATE$, for instance), sharing recovered or dead lists. 

Additional experimentation must be conducted in order to assess its performance on standard $F$ functions and find out the search space shapes in which it can be more effective. 

Some actions must be taken to reduce the size of the infected population after several iterations, that grows exponentially. In this sense, a vaccine must be implemented. This case would involve adding to the recovered list, at a given $VACCINE\_RATE$ healthy individuals. This rate will remain unknown until a vaccine is developed. 

Another suggested research line is using dynamic rates. For instance, the observation of the preliminary effects of the social isolation measures in countries like China, Italy or Spain, suggests that the $INFECT \_RATE$ could be simulated as a Poisson process, but more time and country recoveries is required to confirm this trend.

Finally, for the multi-step forecasting problem analyzed, it would be desirable to generate independent models for each of the values that form the prediction horizon $h$.

\section*{Supplementary material}
Along with this paper, an academic version in Java for a binary codification is provided, with a simple fitness function (\url{https://github.com/DataLabUPO/CVOA_academic}). Additionally, the code in Phyton for the deep learning approach is also provided, with a more complex codification and the suggested implementation, according to the pseudocode provided (\url{https://github.com/DataLabUPO/CVOA_LSTM}).

\section*{Acknowledgments}
The authors would like to thank the Spanish Ministry of Economy and Competitiveness for the support under project TIN2017-88209-C2.

\bibliography{references}
\bibliographystyle{plain}

\end{document}